%% file: main.tex
\documentclass[10pt,twocolumn,letterpaper]{article}

\usepackage{iccv}              


\definecolor{iccvblue}{rgb}{0.21,0.49,0.74}
\usepackage{float}
\usepackage{booktabs}
\usepackage{multirow}
\usepackage{setspace}
\usepackage{lipsum}
\usepackage[pagebackref,breaklinks,colorlinks,allcolors=iccvblue]{hyperref}

\usepackage{makecell}

\usepackage{algorithm}
\usepackage{algorithmic}
\usepackage{amsmath}
\usepackage{graphicx}
\usepackage[accsupp]{axessibility}

\title{Ultra High-Resolution Image Inpainting with Patch-Based \\ Content Consistency Adapter}

\author{Jianhui Zhang$^{1}$ \quad 
    Sheng Cheng$^{2}$ \quad 
    Qirui Sun$^{3}$ \quad 
    Jia Liu$^{1}$ \quad 
    Wang Luyang \quad \\
    Chaoyu Feng \quad 
    Chen Fang$^{3}$ \quad 
    Lei Lei \quad 
    Jue Wang$^{3}$ \quad 
    Shuaicheng Liu$^{1}$\thanks{Corresponding author.} \quad \\
    $^{1}$University of Electronic Science and Technology of China \\ 
    $^{2}$Megvii Technology \quad
    $^{3}$Dzine AI, SeeKoo 
}

\begin{document}

\twocolumn[{%
    \maketitle
    \begin{figure}[H]
    \hsize=\textwidth
    \centering
    \includegraphics[width=2.0\linewidth]{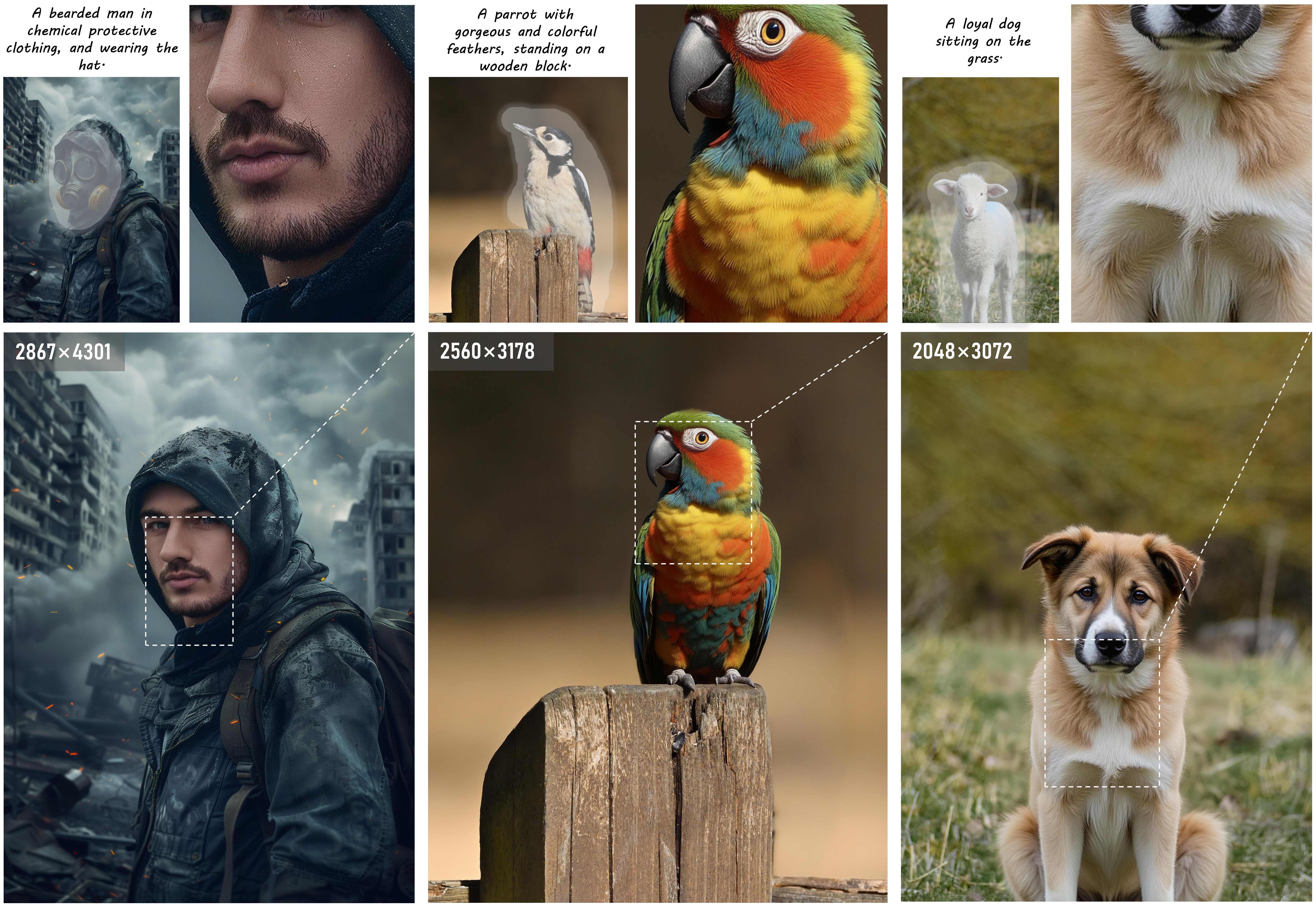}
    \caption{The proposed \textbf{Patch-Adapter}, which enables text-guided high-resolution inpainting at 4K+ resolution while ensuring global content coherence and producing seamlessly blended, visually harmonious results with high fidelity and rich details.}
    \label{fig:teaser}
\end{figure}
}]

\begingroup
\renewcommand\thefootnote{}\footnotetext{
\footnotesize\textsuperscript{*}Corresponding author.
}
\endgroup

\input{0_abstract}    
\input{1_intro}

\input{2_relatedworks}

\input{3_method}

\input{4_experiments}

\input{5_conclusion}

\section*{Acknowledgements}
This work was supported in part by National Natural Science Foundation of China under grant No.62372091 and in part by Hainan Province Key R\&D Program under grant No.ZDYF2024(LALH)001.

\clearpage
{
    \small
    \bibliographystyle{ieeenat_fullname}
    \bibliography{main}
}

\end{document}

%% file: 0_abstract.tex
\begin{abstract}
In this work, we present Patch-Adapter, an effective framework for high-resolution text-guided image inpainting. Unlike existing methods limited to lower resolutions, our approach achieves 4K+ resolution while maintaining precise content consistency and prompt alignment—two critical challenges in image inpainting that intensify with increasing resolution and texture complexity.
Patch-Adapter leverages a two-stage adapter architecture to scale the Diffusion models's resolution from 1K to 4K+ without requiring structural overhauls:
(1)Dual Context Adapter: Learns coherence between masked and unmasked regions at reduced resolutions to establish global structural consistency.
(2)Reference Patch Adapter: Implements a patch-level attention mechanism for full-resolution inpainting, preserving local detail fidelity through adaptive feature fusion.
This dual-stage architecture uniquely addresses the scalability gap in high-resolution inpainting by decoupling global semantics from localized refinement. Experiments demonstrate that Patch-Adapter not only resolves artifacts common in large-scale inpainting but also achieves state-of-the-art performance on the \textit{OpenImages} and \textit{photo-concept-bucket} datasets, outperforming existing methods in both perceptual quality and text-prompt adherence. The code is available at: \url{https://github.com/Roveer/Patch-Based-Adapter}
\end{abstract}

%% file: 1_intro.tex
\section{Introduction}

Image inpainting seeks to restore corrupted images by generating plausible content—a goal that deep learning techniques have dramatically advanced, enabling applications such as virtual try-on~\cite{li2023virtual} and image editing~\cite{huang2024diffusion}. However, most existing methods~\cite{navasardyan2020image, yi2020contextual, yu2018generative} rely solely on the image’s visual context and often overlook high-level semantic guidance from users, a drawback that becomes particularly evident when generating novel content beyond the original scene (e.g., adding a text-specified object).

The emergence of diffusion models~\cite{DDPM,DDIM} has transformed the field, especially through text-guided image completion. This technique allows users to generate new content in designated regions based on textual prompts, supporting tasks like targeted retouching, object replacement or insertion, and modifying attributes such as clothing, color, or expression. Pre-trained diffusion models~\cite{DALLE2, rombach2022high, Imagen} can perform inpainting without fine-tuning; for example, methods like Blended Diffusion~\cite{avrahami2022blended, avrahami2023blended} and DDNM~\cite{wang2022zero} employ masks during diffusion sampling to blend newly generated content with unchanged regions. Nevertheless, a limited understanding of mask boundaries and insufficient contextual integration often lead to incoherent results, particularly during high diffusion timesteps when global scene comprehension is critical.

To address these issues, recent approaches~\cite{ReplaceAnything,rombach2022high,wang2023imagen, xie2023smartbrush, xie2023dreaminpainter, yang2023uni, yu2023inpaint, yang2023magicremover, zhuang2024task} have introduced additional contextual cues and fine-tuned text-to-image models by expanding network inputs. For instance, SDXL-inpainting~\cite{podell2023sdxl} concatenates masks with the original images, which necessitates reinitializing the first convolutional layer to accommodate the modified input. However, such straightforward modifications tend to suffer from suboptimal prompt conditioning and inadequate semantic integration~\cite{SmartBrush,wang2023imagen}. In response, BrushNet~\cite{ju2024brushnet} adds a parallel trainable UNet branch for targeted fine-tuning of pretrained Stable Diffusion(SD) models, while PowerPaint~\cite{zhuang2024task} advances this concept with task-specific architectures that substantially improve textual controllability. More recently, HD-Painter~\cite{manukyan2023hd} attains 2K inpainting resolution via integrated super-resolution processing, achieving a 4× (1024px$\rightarrow$2048px) improvement over conventional methods.

\begin{figure}
    \centering
    \includegraphics[width=0.99\linewidth]{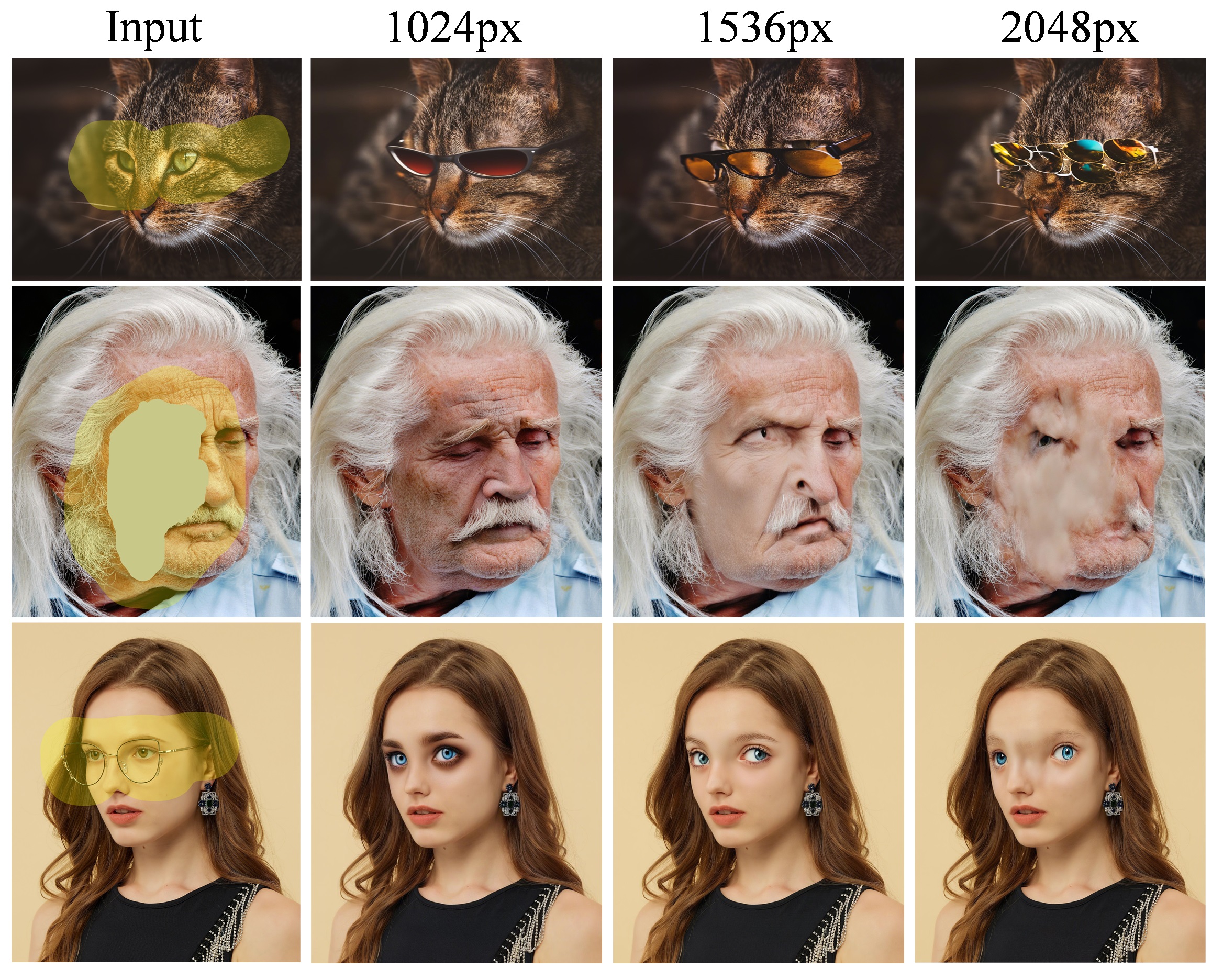}
    \caption{As image resolution increases,  inpainting models tend to produce more artifacts, undermining overall image quality. (The presentation results are from SDXL-inpainting~\cite{podell2023sdxl}.)}
    \label{fig:resolution_increases}
\end{figure}

Despite recent progress, inpainting methods remain challenged by inconsistent completions in filled regions – a flaw that becomes even more pronounced in high-resolution scenarios. We empirically observe that with increasing resolution, these models exhibit diminished attention to unmasked areas and produce inconsistent content (Fig.~\ref{fig:resolution_increases}), which originates from the inherent resolution constraints of pretrained stable diffusion architectures.

In this work, we present Patch-Adapter, a diffusion adaptation framework that actively adapts pretrained SDXL-inpainting models for 4K+ resolution through lightweight parameter grafting, as shown in Fig.~\ref{fig:teaser}. Departing from conventional approaches that passively adapt diffusion models for basic consistency maintenance, we propose dual adaptation strategy that simultaneously: 
\begin{itemize}
    \item Adapts global semantic processing through our Dual Context Adapter (DCA), which actively aligns structural relationships between masked and unmasked region
    \item Adapts local refinement dynamics via Reference Patch Adapter (RPA), implementing context-aware patch adaptation through cross-patch attention.
\end{itemize}
This active adaptation paradigm achieves two critical advancements: (1) Resolution adaptation: Scales pretrained 1K models to \textbf{4K+ regimes while preserving prompt adherence and structural consistency} and (2) Processing adaptation: Transforms standard inpainting workflows into hierarchical patch-aware processing.

To further enhance global coherence, we introduce a hierarchical text prompting mechanism. Global prompts describe the entire scene (e.g., "snow-capped mountains, alpine lakes, and coniferous forests"), while local patch-specific prompts refine regional details (e.g., "texture of pine branches in the foreground"). This dual-level guidance ensures semantic alignment between holistic composition and localized elements.

In summary, our main contributions are as follows:
\begin{itemize}
    \item We propose Patch-Adapter, a parameter-efficient adaptation framework that enables 4K-resolution image inpainting without full-model fine-tuning, maintaining content consistency through learnable parameters grafting.
    \item We introduce a dual-stage adapter framework, comprising a Dual Content Adapter and a Reference Patch Adapter, that effectively extends pre-trained SDXL-inpainting models from 1K to 4K resolution.
    \item We also introduce a hierarchical text prompting mechanism to enhance global coherence and offer an in-depth analysis of each module within the Patch-Adapter.
\end{itemize}

%% file: 2_relatedworks.tex
\section{Related Work}
\label{sec:related}
\subsection{Image Inpainting}
Image inpainting is a long-standing task in computer vision and has been studied for decades. Conventional methods predominantly rely on the CNNs or Transformers architectures~\cite{yang2017high, cao2023generator, zeng2020high, zeng2022aggregated} for restoring masked regions with contextually coherent and visually consistent content~\cite{quan2024deep, xu2023review}. The training of these networks has been facilitated by the adoption of variational auto-encoders ~\cite{peng2021generating, zheng2019pluralistic} and generative adversarial networks ~\cite{liu2021pd, zhao2021large, zheng2022image}. Benefiting from these generative models, the inpainting models can fill in missing or damaged regions in a way that is consistent with the surrounding content, resulting in high-quality inpainted images that are visually coherent and realistic.

Recently diffusion models~\cite{DDPM,DDIM, DALLE2, rombach2022high, Imagen} greatly promoted advancements for image inpainting~\cite{avrahami2022blended, avrahami2023blended,wang2022zero}, where the content is text-guided and controllable. 
Some works aimed to design a training-free approach that can be plug-and-play to any diffusion model. Specifically, Blend Diffusion~\cite{avrahami2022blended, avrahami2023blended} and DDNM~\cite{wang2022zero} strategically designed latent variables and noise during the diffusion model sampling to enhance coherence between generated content in masked regions and the unmasked image. Later, HD-painter~\cite{manukyan2023hd} proposed a prompt-aware attention module that uses the pre-trained weights to increase accuracy. Although the cost of changing the base models is minimal, these methods tend to produce poor results. 

\subsection{Fine-tuning Inpainting Models}
One primary approach in diffusion‐based inpainting fine-tunes pre-trained text-to-image models by conditioning the denoising process on both the inpainting mask and the known region, concatenated with the input latent codes~\cite{ReplaceAnything,rombach2022high,wang2023imagen,xie2023smartbrush,xie2023dreaminpainter,yang2023uni,yu2023inpaint,yang2023magicremover,zhuang2024task}. In contrast, ControlNet-Inpainting~\cite{ControlNet} attached additional parameters to the UNet instead of directly optimizing the base model, employing a parallel encoder architecture that seamlessly integrates its features into a fixed network structure. Subsequently, BrushNet~\cite{ju2024brushnet} leveraged a dual-branch architecture featuring a fully trainable UNet to amplify semantic effects, while PowerPaint~\cite{zhuang2024task} adopted distinct parameters for different completion tasks. More recently, IP-Adapter~\cite{ye2023ip} introduced a learnable attention mechanism that more effectively integrates fine-tuning features by injecting only a few parameters into the attention layers.

Text-guided image inpainting relies on a pre-trained base model, which constrains the resolution to the size of the training images. Consequently, high-resolution inpainting remains underexplored. HD-painter~\cite{manukyan2023hd} pioneers this area by proposing an inpainting-specialized super-resolution model that scales images by 4×, enabling a pipeline for 2048$\times$2048 resolution inpainting. In this work, we introduce the first 4K+ resolution image inpainting capability achieved exclusively through a lightweight adapter-based framework.

%% file: 3_method.tex
\begin{figure*}[t]
    \centering
    \includegraphics[width=0.99\linewidth]{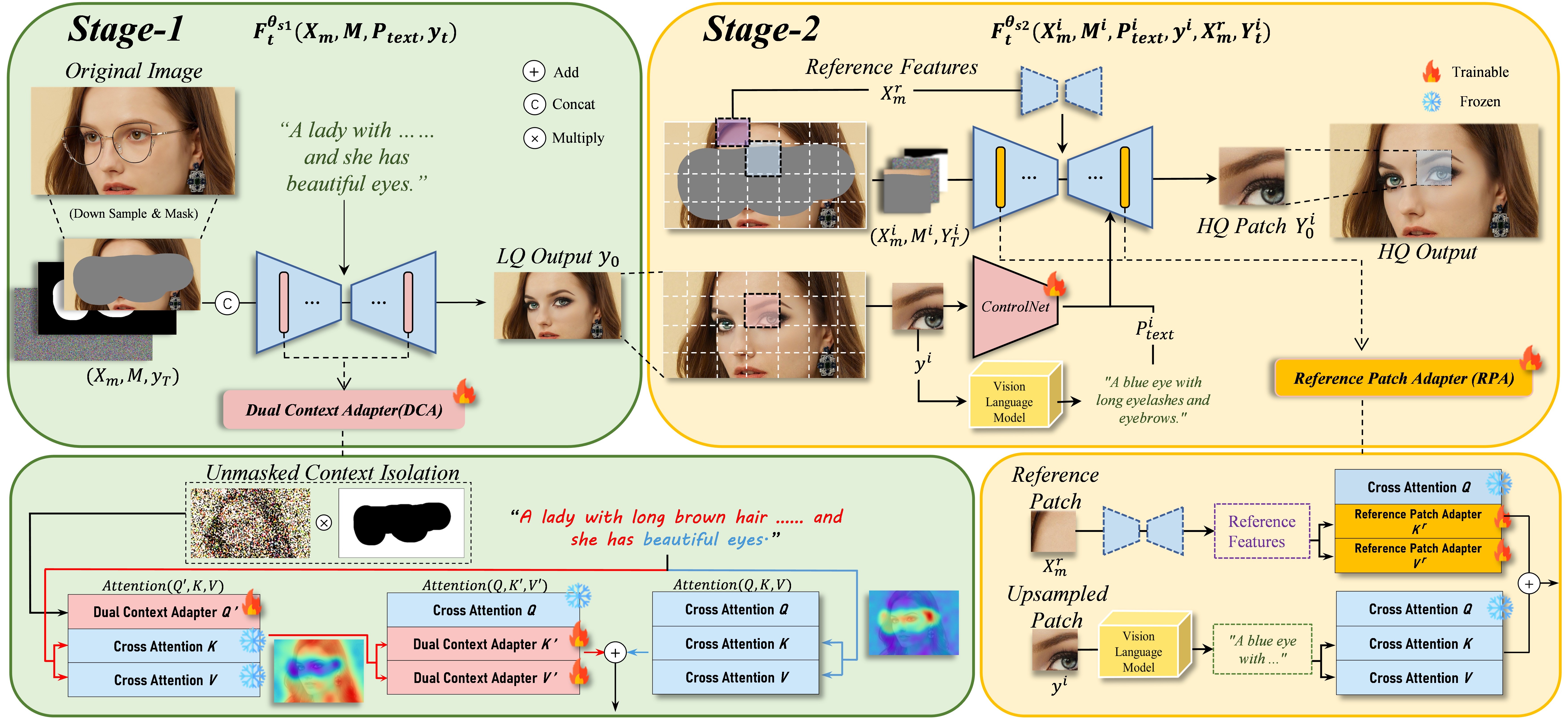}
    \caption{We propose a two-stage pipeline for high-resolution inpainting. Stage 1 leverages a fine-tuned \textbf{Dual Context Adapter} (DCA) to generate visually coherent and contextually accurate content at a lower resolution while balancing both image context and global text prompts. Stage 2 refines the output by utilizing its upsampled result to preserve global structure and employs \textbf{Reference Patch Adapter} (RPA) to capture cross-patch features, enhancing detail richness and fidelity.}
    \label{fig:pipeline_frame}
\end{figure*}

\section{Method}
\label{sec:method}

This section details our two-stage adapter framework (shown in Fig.~\ref{fig:pipeline_frame}) comprising:1) a \textbf{Dual Context Adapter} (DCA) stage for consistent content generation, followed by 2) a \textbf{Reference Patch Adapter} (RPA) stage for high-fidelity detail synthesis. Our pipeline initially conducts base-resolution (1K) inpainting for structural completion (Stage 1). Once Stage 1 is fully trained, Stage 2 builds on the DCA by incorporating RPA to perform high-resolution (4K) patch-based refinement that preserves the original detail fidelity through local context integration.
\subsection{Diffusion-based Inpainting}
 Given a masked image $X_{m}$ and its corresponding mask $\mathcal{M}$, this work proposes to learn a function $\mathcal{F}$ that semantically completes the masked regions under the guidance of a text prompt $\mathcal{P}_{text}$, producing a restored image $y$:
\begin{equation}
    \mathcal{F}:(X_{m}, \mathcal{M}, \mathcal{P}_{\text{text}}) \rightarrow y
\end{equation}

Our approach builds upon the SDXL-inpainting framework~\cite{podell2023sdxl}, a diffusion-based generative model. Following the standard diffusion paradigm, the inpainting process operates iteratively through a $T$-steps Markov chain parameterized by timestep $t$. Formally, our diffusion network $\theta^{\prime}=\{\theta^{\star}, \theta\}$,  which contains the pre-trained fixed weight $\theta^{\star}$ from SDXL-inpainting and the trainable parameter $\theta$, implements the inpainting function at each timestep as:
\begin{equation}
    \mathcal{F}_{t}^{\theta}:(X_{m}, \mathcal{M}, \mathcal{P}_{\text{text}}, y_t) \rightarrow y_{t-1},
\end{equation}
where the $\theta^{\star}$ is omitted for simplicity, $y_T$ is a random noise and $y_0$ is the inpainted image.

\subsection{Stage1: Dual Context Adapter} This stage focuses on resolving the fundamental challenge of preserving \textit{semantic consistency between masked and unmasked regions}. To this end, we design a \textbf{dual-context attention mechanism} that incorporates a Dual Context Adapter (DCA) layer—a parameterized module using \(\theta^{s1}\) to achieve region-adaptive feature modulation. The mechanism is mathematically defined by the governing equation:  
\begin{equation}  
    \mathcal{F}_t^{\theta_{\text{s1}}}:\big(X_m, \mathcal{M}, \mathcal{P}_{\text{text}}, y_t\big) \rightarrow y_{t-1},  
\end{equation}  
where \(y\) represents the intermediate restoration output from this stage, and \(\theta_{\text{s1}}\) denotes the DCA parameter space.  

\subsubsection{Dual Context Adapter (DCA) layer}
Let \( z \in \mathbb{R}^{d} \) denote the visual feature extracted from the masked input tuple \((X_m, \mathcal{M}, y_t)\), and \( c \in \mathbb{R}^{d} \) represent the text feature encoded from the global prompt \( \mathcal{P}_{\text{text}} \). The original SDXL-inpainting attention mechanism computes:  
\begin{equation}  
\mathbf{Z} = \operatorname{Attention}(\mathbf{Q}, \mathbf{K}, \mathbf{V}) = \operatorname{Softmax}\left(\frac{\mathbf{Q K}^\top}{\sqrt{d}}\right)\mathbf{V},
\label{eq:ori_att}
\end{equation}  
where \(\mathbf{Q} = zW^\star_q\), \(\mathbf{K} = cW^\star_k\), \(\mathbf{V} = cW^\star_v\), and \(\{W^\star_q, W^\star_k, W^\star_v\}\) are fixed projection weights from pretrained layers.
For the \( l \)-th attention layer, we introduce learnable parameters \(\theta_{s1}^l = \{W_q^l, W_k^l, W_v^l\}\) to augment the attention computation. The DCA module operates through two key steps:  

\noindent\textbf{Unmasked Context Isolation}:  
   Extract background features via element-wise masking and then generate a background-enhanced query 
   \begin{equation}  
   \mathbf{Q}' = [z \odot (1-M) ] \cdot W_q^l 
   \end{equation}  

\noindent\textbf{Dual-Attention Fusion}:  
   Compute complementary attention maps, \textit{unmask-guided text attention} and \textit{text-refined mask attention}:
   \begin{align}  
   \mathbf{Z}' &= \operatorname{Attention}\big(\mathbf{Q}', \mathbf{K}, \mathbf{V}\big) \\  
   \mathbf{Z}'' &= \operatorname{Attention}\big(\mathbf{Q}, \mathbf{K}', \mathbf{V}'\big)
   \end{align}  
   where \(\mathbf{K}' = \mathbf{Z}' W_k^l\) and \(\mathbf{V}' = \mathbf{Z}' W_v^l\) are text-conditioned projections. Finally, the resulting feature is $\mathbf{Z}_{n} =   \mathbf{Z} + \mathbf{Z}''$.

\subsection{Stage2: Reference Patch Adapter}
\label{sec:rpa}
In this stage, we address the inpainting challenge directly at native image resolution rather than using downsampled versions. To enable inpainting at original resolutions, we implement a patch-wise processing strategy. However, this approach inherently restricts access to unmasked regions from adjacent patches during target patch restoration. Our proposed reference patch adapter mechanism effectively resolves this critical limitation by enabling contextual awareness across patches.

Building upon this notation framework, we formally define the $i$-th patch and its related inputs as $(X_{m}^{i}, \mathcal{M}^{i}, P^{i}_{\text{text}})$, where superscript $i$ preserves spatial indexing across the grid. The current processing stage introduces two critical inputs: (1) the preliminary output $y^{i}$ from Stage 1, and (2) the reference patch $X_{m}^{(r)}$ containing cross-patch contextual information. Our architecture maintains fixed model components including the pretrained SDXL-inpainting backbone and Dual Context Adaptor (DCA) parameters optimized during Stage 1, which are omitted from schematic diagrams for visual clarity while remaining fully operational in implementation:

\begin{equation}  
    \mathcal{F}_t^{\theta_{\text{s2}}}:\big(X_m^{i}, \mathcal{M}^{i}, \mathcal{P}^i_{\text{text}}, y^{i}, X_{m}^{r}, Y^{i}_t\big) \rightarrow Y^{i}_{t-1},  
\end{equation} 
where \(i\) denotes the index of the target patch, \(r\) represents its reference patch index, \(Y_T\) is a standard Gaussian noise input, and \(Y_0\) corresponds to the generated image. Given an image resolution \((H, W)\) and patch dimensions \((n_h, n_w)\), the total patch count is computed as \(N = \frac{H \times W}{n_h \times n_w}\).

\subsubsection{Reference Patch Adapter (RPA) layer}
\noindent\textbf{Reference Patch Selection Strategy.} For each masked patch \( X_m^i \), we dynamically select optimal reference patch \( X_m^r \) by leveraging CLIP~\cite{CLIP} model \(\mathcal{C}\) to compute pairwise cosine similarity across candidate patches. Formally:  
\begin{equation}  
    X^{r}_{m} = \mathop{\arg\max}\limits_{l \neq i} \frac{ \mathcal{C}(y^i)^{\top} \mathcal{C}(X_{m}^l) }{ \| \mathcal{C}(y^i) \|_2 \| \mathcal{C}(X_{m}^l) \|_2 },  
\end{equation}  
where $y^{i}$ denotes the first-stage output patch associated with $X_m^i $ and the constraint \( l \neq i \) ensures exclusion of self-reference.  

\noindent\textbf{Reference Adapter Module.} Upon selecting the reference patch, we extract the reference feature $z^{r} \in \mathbb{R}^d$ by propagating the triplet $(X^{r}_m, \mathcal{M}, \mathcal{P}_{\text{text}}^{j})$ through the Stage 1-trained U-Net, where the feature is derived from the attention layers' outputs. The reference adapter incorporates two trainable parameters per layer: $\theta^{l}_{s2}=W_{k}^{l},W_{v}^{l}$ for the $l$-th transformer layer. Given preliminary feature $\mathbf{Q}$ from Eq.~\ref{eq:ori_att}, the adaptation process is formally defined as:

\begin{align}  
   \mathbf{K}^{r} &= z^{r} W_k^l \quad \text{(ref-conditioned key projection)} \\  
   \mathbf{V}^{r} &= z^{r} W_v^l \quad \text{(ref-conditioned value projection)} \\
   \mathbf{Z}^{r} &= \operatorname{Attention}\big(\mathbf{Q}, \mathbf{K}^{r}, \mathbf{V}^{r}\big) \\  
   \mathbf{Z}^{r}_{\text{n}} &= \mathbf{Z}^{r} + \mathbf{Z}
\end{align}

This residual architecture progressively integrates reference-aware adaptations through additive feature composition.

\subsection{Technical Enhancements}
We provide a detailed description of the proposed hierarchical text prompting mechanism, along with several techniques commonly employed in diffusion models.

\noindent\textbf{Stage 1 \& Stage 2:hierarchical text prompting}
To further enhance global coherence, we propose a hierarchical text prompting mechanism that provides dual-level guidance for semantic alignment. At the global level, scene-wide prompts (e.g., "snow-capped mountains, alpine lakes, and coniferous forests") describe the entire image, while local patch-specific prompts (e.g., "texture of pine branches in the foreground") refine regional details. This combination ensures consistent composition between holistic semantics and localized elements.

To improve inter-patch consistency, we refine patch-specific prompts by leveraging Vision-Language Models (VLM)~\cite{xiao2024florence}. The patch-wise outputs from Stage 1 are batch-processed through the VLM framework to generate context-aware textual descriptors for all patches simultaneously. Unlike conventional methods that rely on primitive mask-derived prompts (e.g., "object removal"), our approach formulates prompts as a combination of foreground descriptions from the masked region and background scene context from the known region, expressed as:
\begin{equation}
\mathcal{P}_{\text{text}}^{g} = \text{Foreground}(\mathcal{M}) + \text{Background}(1 - \mathcal{M})
\end{equation}
This context-aware prompting strategy enables more accurate and semantically consistent inpainting across patches.

\noindent\textbf{Stage 2:ControlNet}
We adopt ControlNet to guide high-resolution patch refinement by effectively modulating global low-frequency components. It injects structural guidance without altering the pre-trained base model, preserving global consistency. Specifically, ControlNet extracts discriminative features from each patch 
$y^{i}$ that encode both structural and semantic cues, serving as explicit control signals to maintain local coherence and overall scene alignment.

\noindent\textbf{Stage 2:Blended Diffusion}
At each timestep $t$, given the intermediate output $Y^{i}_{t-1}$ and the masked input image $X^{i}_{m}$, we first simulate the inpainting process by diffusing $X^{i}_{m}$ with Gaussian noise over $T$ timesteps to obtain $Y^{i}_{m, t-1}$. The blended feature map is then computed through a mask-guided fusion:
\begin{equation}
    Y^{i}_{t-1} = Y^{i}_{t-1} \odot \mathcal{M}^{i} + Y^{i}_{m, t-1} \odot (1 - \mathcal{M}^{i})
\end{equation}
where $M^{i}$ denotes the binary mask, and $\odot$ represents element-wise multiplication. This operation preserves known regions from $Y^{i}_{t-1}$ while integrating inpainted content from $Y^{i}_{m, t-1}$ in masked areas.

%% file: 4_experiments.tex
\begin{figure*}[t]
    \centering
        \includegraphics[width=0.99\linewidth]{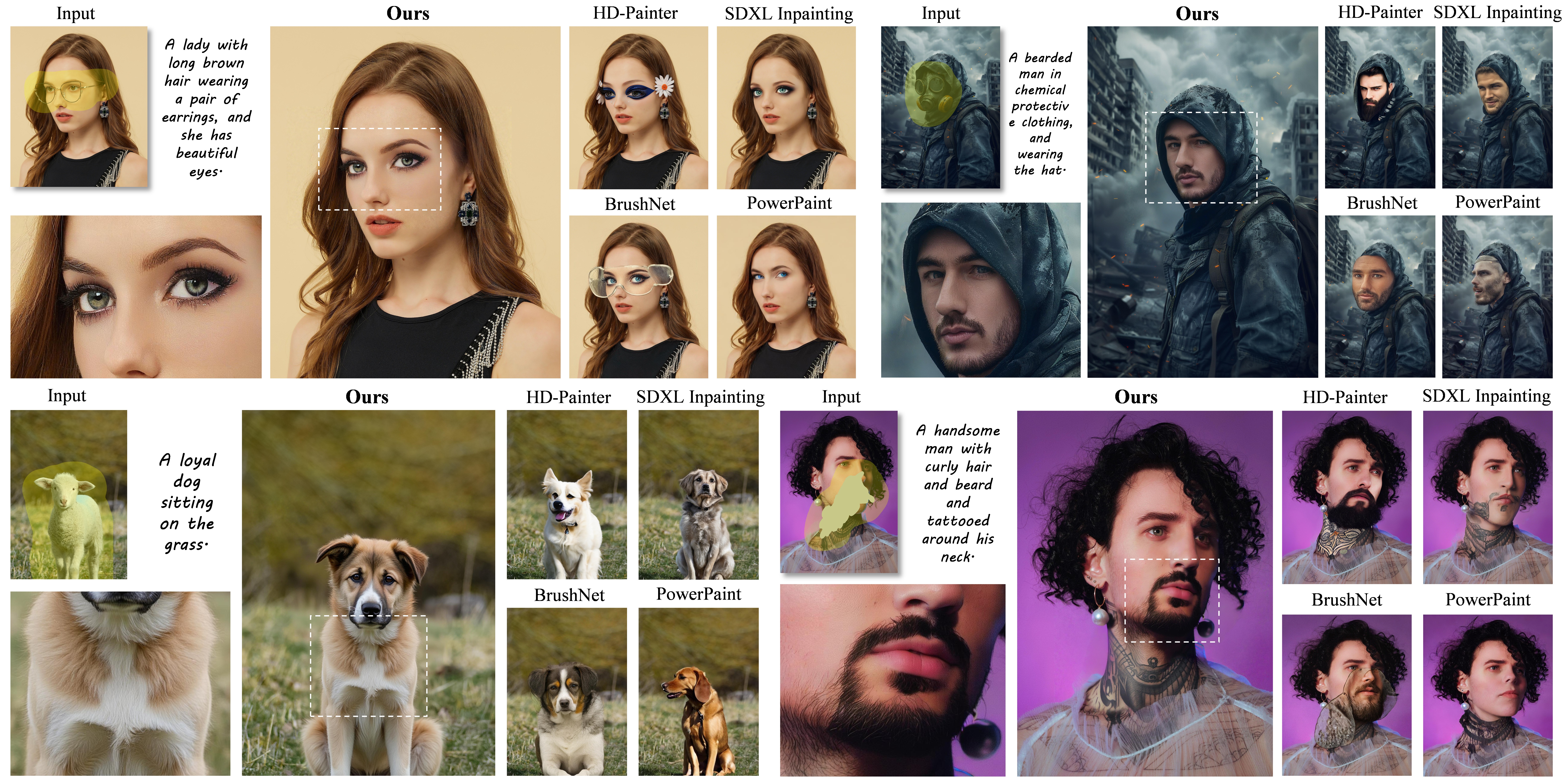}
        \caption{A qualitative evaluation comparing our proposed pipeline with existing methods. Our approach achieved state‐of‐the‐art performance in content accuracy, visual aesthetics, texture consistency both inside and outside the mask.Furthermore, unlike other models that inadvertently introduce image degradation and blur, our approach generate exceptionally realistic, meticulously defined details in high‐resolution images. }

    \label{fig:ppl_compare}

    \vspace{-5mm}
\end{figure*}

\input{main_table}

\section{Experiments}

\subsection{Implementation Details}

\noindent\textbf{Datasets.}
For benchmarking purpose, we evaluate the proposed method through experiments conducted on \textit{OpenImages} and \textit{photo-concept-bucket} datasets.

For Stage 1, the training data includes 211,688 images from OpenImage~\cite{kuznetsova2020open}, each annotated with comprehensive text descriptions. We generate masks for 60\% of the training images using simulated brush strokes (via BrushNet~\cite{ju2024brushnet}), regular geometric shapes, or random shape combinations. The remaining 40\% use segmentation-based masks. For evaluation, we select 5,000 images excluded from training: half are masked randomly, and half use segmentation-based masking to match the training approach. This setup aligns with HD-Painter~\cite{manukyan2023hd} and PowerPaint~\cite{zhuang2024task}.

For Stage 2, training and evaluation use the photo-concept-bucket dataset, with 2,000 high-resolution images for benchmarking. This dataset challenges the model to generate realistic scenes and coherent inpainting results.

\begin{figure*}[t]
    \centering
    \includegraphics[width=0.97\linewidth]{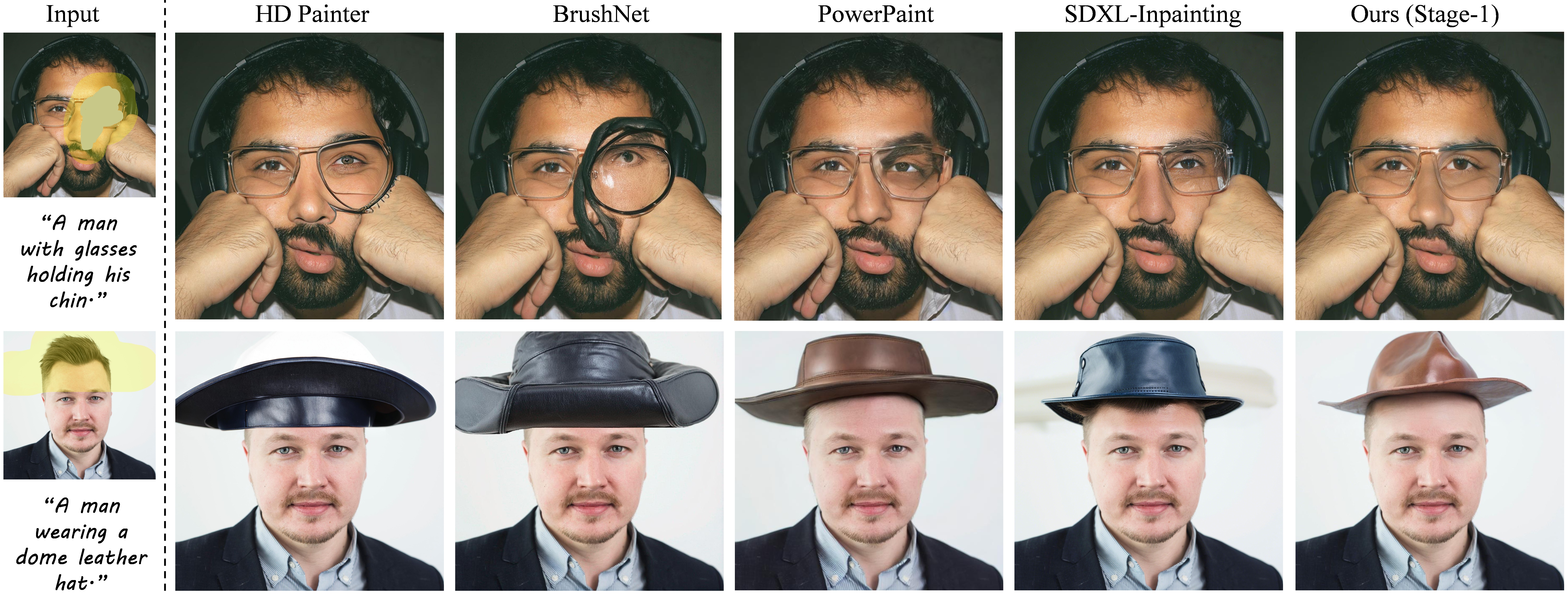}
    \caption{In our qualitative comparison, the model incorporating Dual Context Adapter (DCA) demonstrates superior performance in challenging scenarios, effectively handling tasks such as object removal, image restoration, and object insertion. In contrast, other methods often struggle with contextual understanding, leading to unpredictable color discrepancies and content artifacts.}
    \label{fig:stage1_compare}
\end{figure*}
\begin{figure*}[t]
    \centering
    \includegraphics[width=0.97\linewidth]{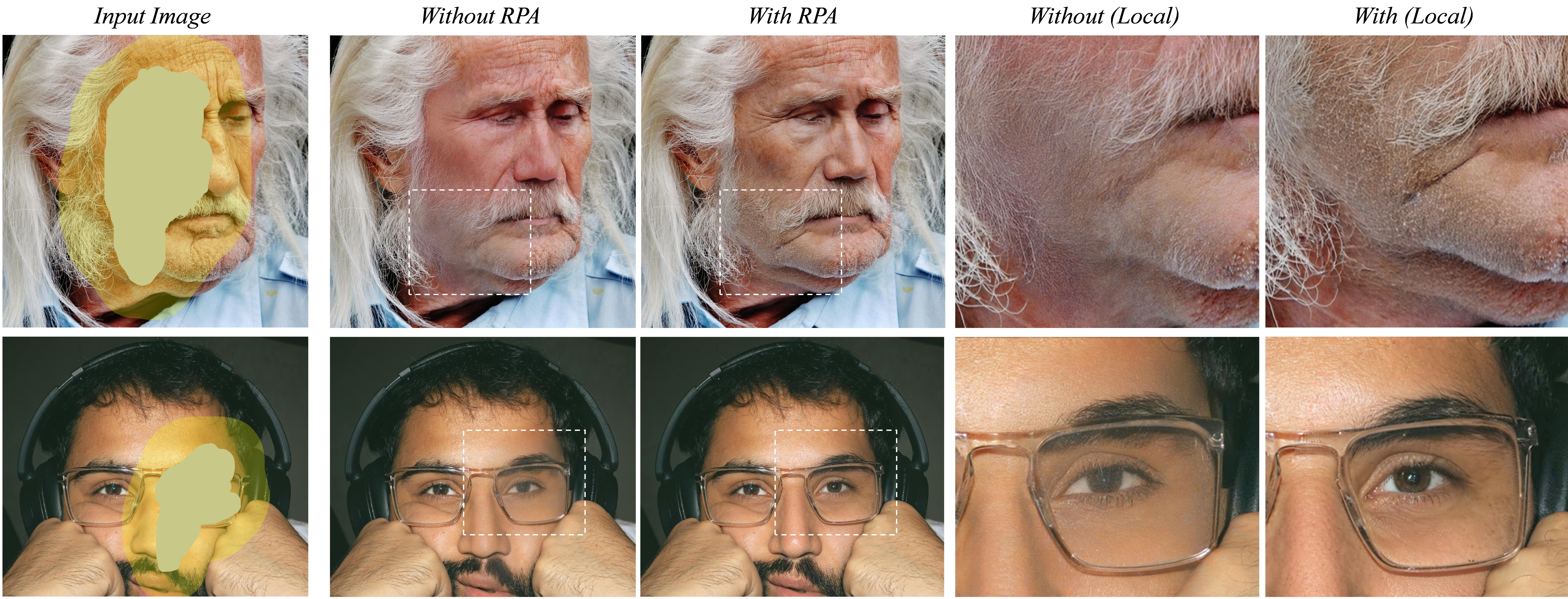}
    \caption{Ablation study of fine-tuning RPA in stage 2. Models incorporating the RPA exhibit enhanced texture consistency in image generation tasks.}
    \label{fig:stage2_ablation}
    \vspace{-4mm}
\end{figure*}

\noindent\textbf{Training and Inference.}
The model was trained in two distinct stages, Stage 1 involved fine-tuning the Dual Context Adapter while Stage 2 fine-tuned both the proposed Reference Patch Adapter (RPA) and the ControlNet~\cite{ControlNet}.

In Stage 2, given that SDXL-inpainting~\cite{podell2023sdxl} achieves optimal performance at a resolution of 1024×1024, all high-resolution images were cropped to 2048×2048, then split into four equal parts. Two segments were randomly selected to serve as inputs: one for the LQ (Low Quality) input and one for the reference patch input, both set at a resolution of 1024×1024. Random masking was employed in a manner consistent with Stage 1 to generate the requisite masks, while image degradation was simulated following the setting used by Real-ESRGAN~\cite{wang2018esrgan}.

For the training procedure, we employed the AdamW optimizer with a learning rate of 0.00002 and a batch size of 128, utilizing Nvidia A6000 GPUs. 

The inference process was carried out using the EulerDiscreteScheduler~\cite{karras2022elucidating}, with a total of 30 inference steps and classifier-free guidance (CFG)~\cite{ho2021classifier} scale of 7.0.

\subsection{Comparison with Existing Methods}

\textbf{Baseline.}
To comprehensively evaluate the effectiveness of our proposed method, we conducted comparisons with state-of-the-art approaches in the field of image inpainting, including PowerPaint~\cite{zhuang2024task}, BrushNet~\cite{ju2024brushnet}, HD-Painter~\cite{manukyan2023hd}, and SDXL-inpainting~\cite{podell2023sdxl}. Notably, SDXL-inpainting is a fine-tuned model based on the open-source SDXL~\cite{podell2023sdxl} framework. 
In our research, we utilized SDXL-inpainting as the foundational model, introducing novel enhancements to improve its contextual understanding capabilities.

\noindent\textbf{Evaluation Metrics.}
Following standard evaluation practices, we adopt four widely used metrics to quantitatively assess inpainting performance: Fréchet Inception Distance (FID)~\cite{heusel2017gans}, CLIP Score~\cite{CLIP}, LPIPS~\cite{zhang2018unreasonable}, and Aesthetic Score~\cite{schuhmann2022laion}. 
Specifically, FID is used to measure the perceptual quality of the inpainted images, while the CLIP Score quantifies the semantic alignment between the generated content and the given text prompt. LPIPS is utilized to assess reconstruction consistency, ensuring structural coherence with the original image. 
In addition, we incorporate an Aesthetic Score to assess the overall aesthetic quality of the generated images, providing a comprehensive evaluation of the inpainting results.

\begin{figure}[t]
    \centering
    \includegraphics[width=1.0\linewidth]{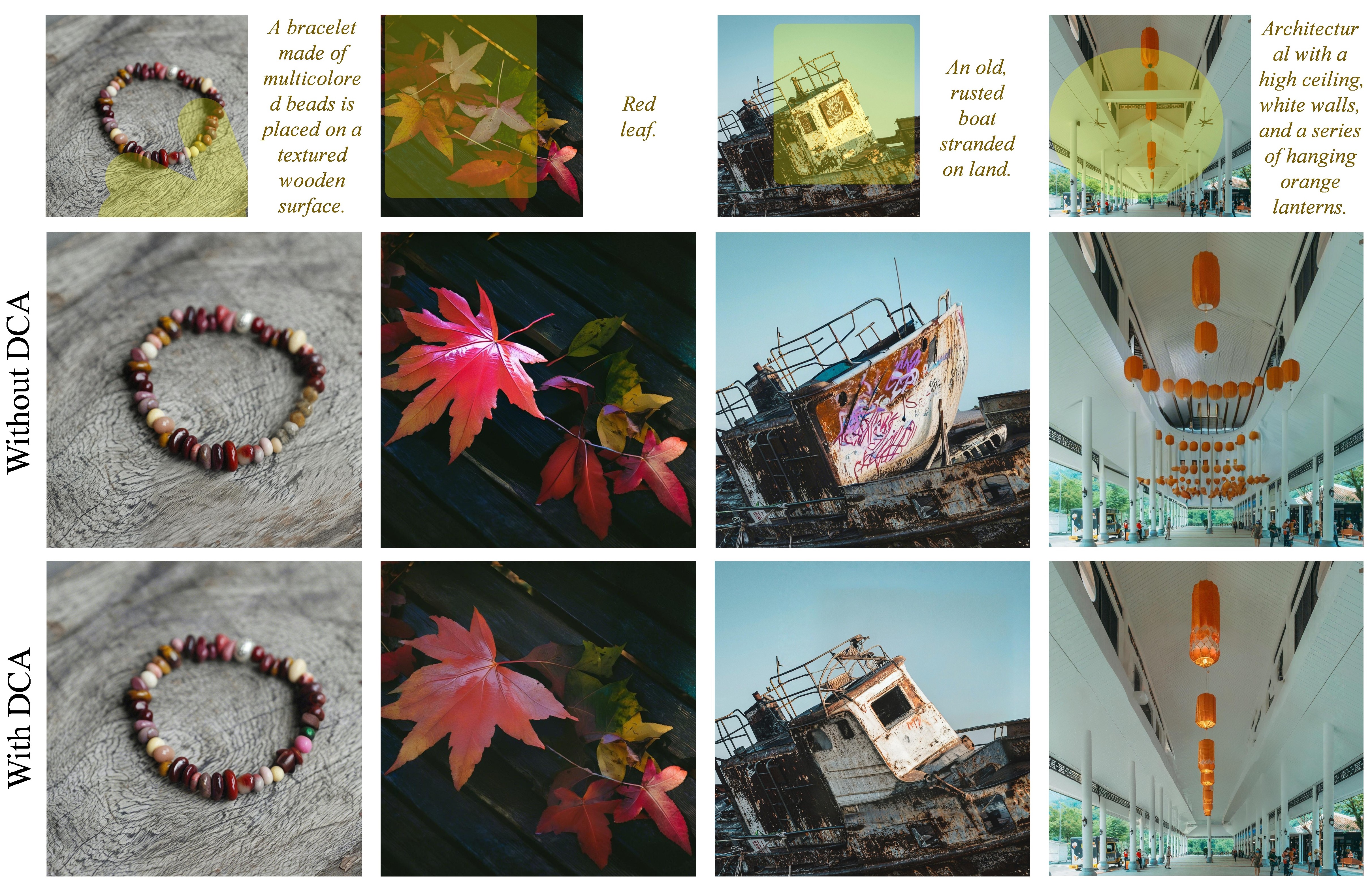}
    \caption{Ablation study of DCA in stage 1. Models incorporating DCA demonstrate superior performance in semantic accuracy, content coherence, and seamless integration.}
    \label{fig:stage1_ablation}
    \vspace{-2mm}
\end{figure}

\input{stage1_table}

\noindent\textbf{Quantitative Comparisons and Qualitative Comparisons.}
To evaluate the effectiveness of the proposed method, given that the best performance of previous methods is achieved at the resolution 1K, we first conducted quantitative evaluations of the Stage 1 fine-tuning. As shown in Tab.~\ref{tab:Stage1}, fine-tuning DCA enhances the model's contextual understanding, effectively leveraging global prompt information and utilizing contextual cues even when provided with local prompts.
As depicted in Fig.~\ref{fig:stage1_compare}, the subjective results demonstrate that our Stage 1 fine-tuning of DCA enables the model to effectively comprehend contextual information, actively guiding the generation process to achieve seamless and high-quality inpainting. In contrast to other models that produce unpredictable content degradation due to inadequate context comprehension, our model exhibits superior performance across various text-guided tasks.

We evaluated our full pipeline on a high-resolution real-world dataset. As shown in Tab.~\ref{tab:metrics}, 
the highest Aesthetic Score (6.021) and CLIP Score (26.806), reflecting its ability to generate visually pleasing and semantically aligned content. Moreover, the notably low LPIPS (0.153) corroborates the artifact-free nature of our inpainted regions, indicating that they blend seamlessly with the original image. Collectively, our model excels in high-resolution inpainting by ensuring consistency, accurately aligning with text prompts, and producing high-quality, artifact-free results.

As illustrated by the qualitative assessments in Fig.~\ref{fig:ppl_compare}, our approach substantially improves the correctness of inpainted content compared to other methods, which often produce seams, contextually irrelevant elements, or even completely corrupted regions. Furthermore, our model preserves the high-resolution characteristics of the original images by synthesizing exquisitely refined and meticulously delineated details, thereby ensuring both visual fidelity and coherence.

\subsection{Ablation Study}

\textbf{Dual Context Adapter (DCA).}
We compare the original SDXL-inpainting model with our variant incorporating Dual Context Adapter (DCA). As shown in the last two rows of Tab.~\ref{tab:Stage1}, our fine-tuning strategy achieves better text alignment and overall image quality, demonstrating the benefit of contextual adaptation. As illustrated in Figure \ref{fig:stage1_ablation}, subjective evaluations further demonstrate that our model successfully resolves issues present in the original, including content inconsistencies, improper stitching, and style discrepancies.

\input{stage2_table}

\noindent\textbf{Reference Patch Adapter (RPA).}
Furthermore, Tab.~\ref{tab:RPA_Ablation} reports quantitative results for models with and without the Reference Patch Adapter (RPA). It is evident that incorporating cross-patch reference information markedly enhances both the aesthetic appeal and reconstruction consistency of the generated images. As shown in Fig.~\ref{fig:stage2_ablation}, RPA enables accurate texture transfer from reference patches to inpainted regions, which is especially effective for portrait restoration.

%% file: main_table.tex
\begin{table}[t]
    \centering
    \small
    \resizebox{1\columnwidth}{!}{
        \begin{tabular}{lcccc}
            \hline
            \textbf{Model Name} 
            & \textbf{\makecell{FID}} $\downarrow$ 
            & \textbf{\makecell{Aesthetic score}} $\uparrow$ 
            & \textbf{\makecell{CLIP Score}} $\uparrow$ 
            & \textbf{\makecell{LPIPS}} $\downarrow$ \\
            \hline
            BrushNet   \cite{ju2024brushnet}       & 36.334 & 5.146  & 26.212 & 0.307  \\ 
            PowerPaint  \cite{zhuang2024task}      & 23.652 & 5.712  & 26.646 & 0.308  \\
            HD-Painter \cite{manukyan2023hd}       & 22.262 & 5.949  & 26.576 & 0.230  \\
            SDXL Inpainting \cite{podell2023sdxl}  & 17.660 & 6.012  & 26.771 & 0.208  \\
            \hline
            \textbf{Ours}               & \textbf{14.594} & \textbf{6.021}  & \textbf{26.806} & \textbf{0.153}  \\
            \hline
        \end{tabular}
    }
    
    \caption{Quantitative comparison for high-resolution inpainting on 2,000 high-resolution images.}
    \label{tab:metrics}
    \vspace{-6mm}
\end{table}

%% file: stage1_table.tex
\begin{table}
    \centering
    \small
    \resizebox{\columnwidth}{!}{
        \begin{tabular}{lcccc}
            \toprule
            \multicolumn{1}{c}{} 
            & \multicolumn{4}{c}{\textbf{Random Mask and Global Prompt}} \\  
            \textbf{Model Name}
            & \textbf{\makecell{FID}} $\downarrow$ 
            & \textbf{\makecell{Aesthetic score}} $\uparrow$ 
            & \textbf{\makecell{CLIP Score}} $\uparrow$ 
            & \textbf{\makecell{LPIPS}} $\downarrow$ \\

            \midrule
            BrushNet   \cite{ju2024brushnet}       & 31.853 & 4.683  & 26.199 & 0.152  \\ 
            PowerPaint  \cite{zhuang2024task}      & 19.661 & 5.471  & 25.974 & 0.179  \\
            HD-Painter \cite{manukyan2023hd}       & 25.111 & 5.348  & 26.381 & 0.150  \\
            \textbf{SDXL Inpainting} \cite{podell2023sdxl}  & 13.326 & 5.480  & 26.268 & 0.129  \\
            \midrule
            \textbf{Ours}                        & \textbf{12.167} & \textbf{5.591}  & \textbf{26.458} & \textbf{0.128}  \\
            \bottomrule
            \multicolumn{1}{c}{} 
            & \multicolumn{4}{c}{\textbf{Segmentation Masks and Local Prompt}} \\ 
            \textbf{Model Name}
            & \textbf{\makecell{FID}} $\downarrow$ 
            & \textbf{\makecell{Aesthetic score}} $\uparrow$ 
            & \textbf{\makecell{CLIP Score}} $\uparrow$ 
            & \textbf{\makecell{LPIPS}} $\downarrow$ \\

            \midrule
            BrushNet   \cite{ju2024brushnet}       & 16.211 & 5.058  & 26.735 & 0.105  \\ 
            PowerPaint  \cite{zhuang2024task}      & 12.481 & 5.543  & 26.865 & 0.120  \\
            HD-Painter \cite{manukyan2023hd}       & 11.694 & 5.541  & 26.712 & 0.097  \\
            \textbf{SDXL Inpainting} \cite{podell2023sdxl}  & 9.565  & 5.559  & 26.990 & 0.092  \\
            \midrule
            \textbf{Ours}                        & \textbf{9.427} & \textbf{5.598}  & \textbf{27.002} & \textbf{0.089}  \\
            \bottomrule
        \end{tabular}
    }
    \vspace{-2mm}
    \caption{Quantitative evaluation of two mask and prompt input methods on Openimage, The rows labeled \textbf{SDXL Inpainting} and \textbf{Ours} represent an ablation study, contrasting the performance with and without Dual Context Adapter (DCA).}
    \label{tab:Stage1}
    \vspace{-4mm}
\end{table}

%% file: stage2_table.tex
\begin{table}
    \centering
    \small
    \resizebox{\columnwidth}{!}{
        \begin{tabular}{lcccc}
            \hline
            \textbf{} 
            & \textbf{\makecell{FID}} $\downarrow$ 
            & \textbf{\makecell{Aesthetic score}} $\uparrow$ 
            & \textbf{\makecell{CLIP Score}} $\uparrow$ 
            & \textbf{\makecell{LPIPS}} $\downarrow$ \\
            \hline
            With \textbf{RPA}         & 14.594 & 6.021  & 26.8063 & 0.153  \\ 
            Without \textbf{RPA}      & 16.144 & 5.910  & 26.8062 & 0.161  \\
            \hline
        \end{tabular}
    }
    \vspace{-2mm}
    \caption{Ablation study of Reference Patch Adapter(RPA).}
    \label{tab:RPA_Ablation}
    \vspace{-4mm}
\end{table}

%% file: 5_conclusion.tex
\section{Conclusion}
In this work, we address the critical challenge of text-guided high-resolution (4K) image inpainting, a task where existing methods primarily rely on fine-tuning 1K-pretrained diffusion models—a strategy that struggles to scale effectively. Departing from parameter-intensive adaptation paradigms, we propose an innovative dual-stage adapter-based architecture that uniquely enables patch-wise processing while maintaining cross-patch content consistency. Extensive experiments demonstrate that our method not only retains the compositional reasoning capabilities of 1K-scale diffusion priors but also enables pixel-accurate 4K+ inpainting. This work establishes a new pathway for deploying lightweight, resolution-agnostic inpainting systems without compromising computational sustainability.

\noindent\textbf{Limitation.} While our method demonstrates superior performance, the patch-based approach introduces a slight increase in inference time. Future work will focus on improving the efficiency of inference.